%% file: main.tex
\documentclass[10pt,twocolumn,letterpaper]{article}

\usepackage{cvpr}              
\usepackage{multirow} 
\usepackage{graphicx}
\usepackage{array}

\definecolor{cvprblue}{rgb}{0.21,0.49,0.74}
\usepackage[pagebackref,breaklinks,colorlinks,allcolors=cvprblue]{hyperref}
\usepackage[accsupp]{axessibility}  

\def\paperID{110} 
\def\confName{CVPR}
\def\confYear{2025}

\title{Distillation-Supervised Convolutional Low-Rank Adaptation for Efficient Image Super-Resolution}

\author{
    Xinning Chai$^{1}$ \quad 
    Yao Zhang$^{1}$ \quad 
    Yuxuan Zhang$^{1}$ \quad 
    Zhengxue Cheng$^{1}$\thanks{ Corresponding author} \\ 
    Yingsheng Qin$^{2}$ \quad
    Yucai Yang$^{2}$ \quad 
    Li Song$^{1}$ \\
    $^1$Shanghai Jiao Tong University, China \quad $^2$Transsion, China \\
    {\tt\small \{chaixinning, yao\_zhang, 67keudyhsi, zxcheng, song\_li\}@sjtu.edu.cn} \\
    {\tt\small \{yingsheng.qin, yucai.yang\}@transsion.com}
}

\begin{document}
\maketitle
\input{sec/0_abstract}

\input{sec/1_intro}

\input{sec/2_related_work}

\input{sec/3_method}

\input{sec/4_Experiments}

\input{sec/5_Conclusion}
\input{sec/6_acknowledgment}
{
    \small
    \bibliographystyle{ieeenat_fullname}
    \bibliography{main}
}


\end{document}

%% file: sec/0_abstract.tex
\begin{abstract}
Convolutional neural networks (CNNs) have been widely used in efficient image super-resolution. However, for CNN-based methods, performance gains often require deeper networks and larger feature maps, which increase complexity and inference costs.
Inspired by LoRA's \cite{lora1} success in fine-tuning large language models, we explore its application to lightweight models and propose Distillation-Supervised Convolutional Low-Rank Adaptation (DSCLoRA), which improves model performance without increasing architectural complexity or inference costs.
Specifically, we integrate ConvLoRA \cite{aleem2024convlora} into the efficient SR network SPAN \cite{wan2024swift} by replacing the SPAB module with the proposed SConvLB module and incorporating ConvLoRA layers into both the pixel shuffle block and its preceding convolutional layer.
DSCLoRA leverages low-rank decomposition for parameter updates and employs a spatial feature affinity-based knowledge distillation strategy \cite{he2020fakd} to transfer second-order statistical information from teacher models (pre-trained SPAN) to student models (ours).
This method preserves the core knowledge of lightweight models and facilitates optimal solution discovery under certain conditions.
Experiments on benchmark datasets show that DSCLoRA improves PSNR and SSIM over SPAN while maintaining its efficiency and competitive image quality.
Notably, DSCLoRA ranked first in the Overall Performance Track of the NTIRE 2025 Efficient Super-Resolution Challenge \cite{ren2025tenth}. Our code and models are made publicly available at \url{https://github.com/Yaozzz666/DSCF-SR}.

\end{abstract}

%% file: sec/1_intro.tex
\section{Introduction}
\label{sec:intro}

\begin{figure}[t]
  \centering
  \includegraphics[width=1\linewidth]{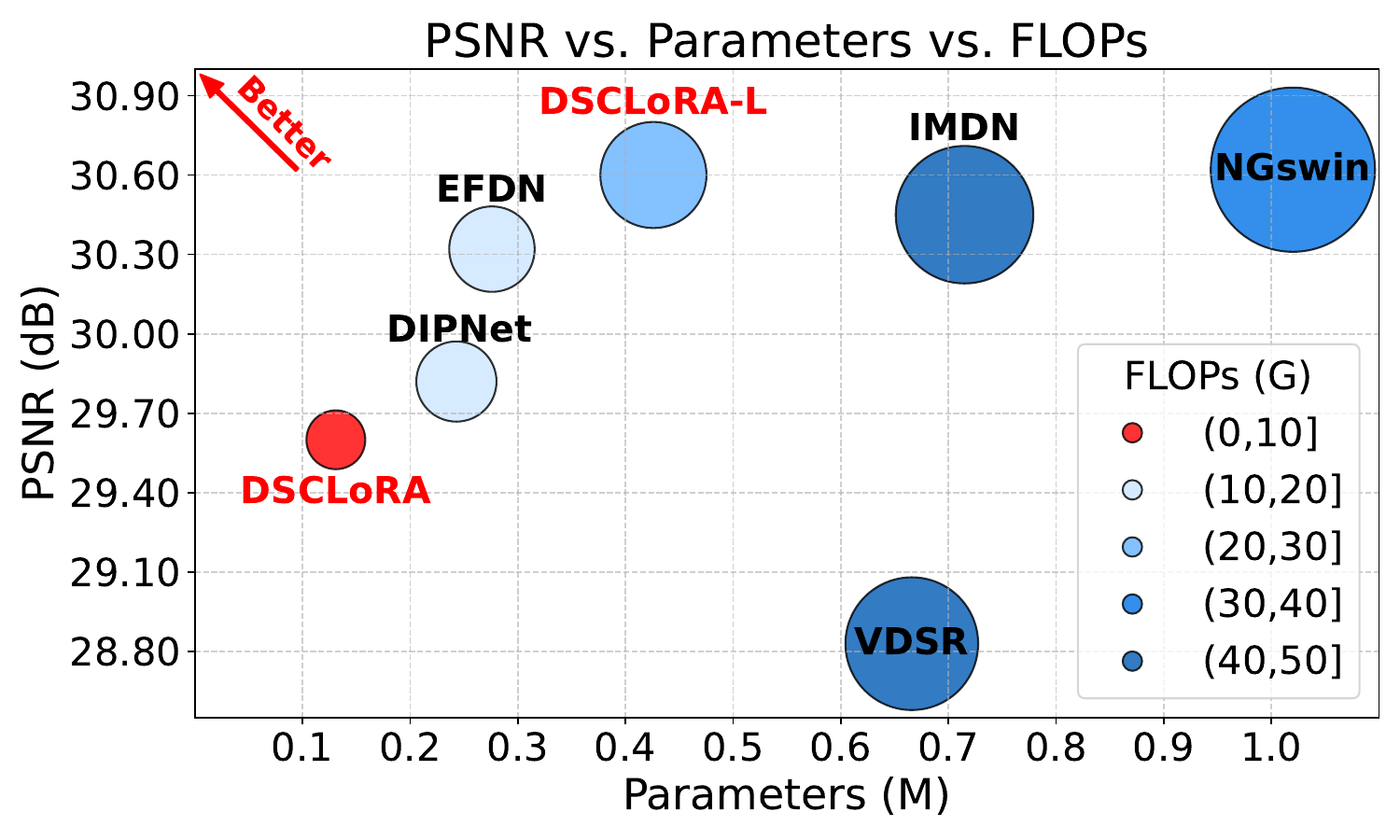}
  \caption{Comparison of Parameters, FLOPs, and PSNR for models on the Manga109 dataset in the $\times4$ scale SR task. DSCLoRA is our 26-channel model while DSCLoRA-L is 48-channel. The color red indicates that the flops are the smallest. The size of the circle represents the number of parameters, the closer to the upper left the better the model is.}
  \label{fig:comparison}
\end{figure}

Single Image Super-Resolution (SISR) is a key technique aimed at reconstructing high-resolution (HR) images from corresponding low-resolution (LR) inputs. It has been widely applied in areas such as medical image analysis, remote sensing, video surveillance, and other fields requiring detailed visual information \cite{do1,do2,do3,do4,do5}.
Advancements in deep learning have significantly improved SISR performance with a wide range of deep neural networks, including convolutional neural networks (CNNs), residual networks, generative adversarial networks (GANs), and transformer-based models \cite{srcnn,fsrcnn,vdsr,drcn,carn,ecbsr,lap,rfdn,imdn,safmn,efdn}. 

To enhance the performance of existing CNN-based SR models, most prior approaches focus on optimizing network design, often by deepening convolutional layers, which increases computational overhead. Our objective, however, is to improve metrics like PSNR without adding additional burden. 
LoRA (Low-Rank Adaptation) \cite{lora1} is a parameter-efficient fine-tuning technology that was often used for fine-tuning large language models in the past. However, for lightweight and efficient super-resolution models, LoRA can also be used for efficient fine-tuning. This approach reduces computational costs and memory requirements, achieves fast convergence, and maintains the basic knowledge of the model, alleviating catastrophic forgetting and making it easier to find the optimal solution \cite{aleem2024convlora}.
Specifically, we leverage ConvLoRA \cite{aleem2024convlora}, which applies LoRA to convolutional layers. By freezing the original model weights and fine-tuning with ConvLoRA, we aim to achieve performance gains. Nonetheless, to address the limited improvements from fine-tuning alone, we further adopt knowledge distillation, a cost-effective technique for boosting lightweight model performance. For effective distillation, the key is to find an appropriate mimicry loss function that can successfully propagate valuable information to guide the training process of the student model. Drawing on the superiority of spatial affinity-based loss in FAKD \cite{he2020fakd}, we incorporate spatial affinity-based loss during the distillation process to enhance the SR model's performance without additional computational complexity.
\setlength{\parskip}{0pt} 

Through the above analysis, we extend LoRA's application to image super-resolution by leveraging SPAN \cite{wan2024swift} as a foundation and proposing a novel framework: the Distillation-Supervised Convolutional Low-Rank Adaptation model (DSCLoRA model). Specially, we propose the SConvLB module, which integrates LoRA layers into pre-trained convolutional layers. Additionally, DSCLoRA is applied to the pixel rearrangement module and its preceding convolutional layer, forming our final model.

In summary, our main contributions are as follows:
\setlength{\parskip}{0pt} 
\begin{itemize}
\item We introduce ConvLoRA \cite{aleem2024convlora} into the efficient single-image super-resolution (SISR) task and propose a novel method, termed \textbf{DSCLoRA}, which can be integrated into existing models for fine-tuning, achieving performance improvement without additional parameters or computational overhead.
\item We employ hybrid distillation training, combining spatial affinity distillation \cite{he2020fakd} for second-order statistics transfer, pixel-level distillation loss, and L1/L2 reconstruction losses, optimized via a unified loss function.
\item The improved performance of the SISR model, driven by DSCLoRA model and its advantage of incurring no additional cost, is demonstrated across several benchmark datasets. Additionally, the effectiveness of using spatial affinity-based distillation is experimentally validated.
\end{itemize}


%% file: sec/2_related_work.tex
\section{Related Work}
\label{sec:2}
\subsection{Efficient Image Super Resolution}
Previous efficient super-resolution models can be mainly divided into the following categories:
Basic convolutional neural network (CNN),
Recursive and deep structure,
Lightweight residual and module design and
Attention mechanism and aggregation network.
SRCNN \cite{srcnn} proposes a deep learning method for single image super-resolution that directly learns an end-to-end mapping between low/high resolution images. The mapping is represented as a deep convolutional neural network that takes a low-resolution image as input and outputs a high-resolution image. SRCNN has shown superior performance to previous hand-crafted models in terms of both speed and restoration quality. However, the high computational cost still hinders its practical application. Real-time performance (24 fps) is required. FSRCNN \cite{fsrcnn} proposes a compact hourglass-shaped CNN structure to achieve faster and better SR. VDSR \cite{vdsr} uses a very deep convolutional network inspired by VGG-net used for ImageNet classification \cite{vgg}. By cascading small filters many times in a deep network structure, contextual information over large image regions is exploited in an efficient way. VDSR learn residuals only and use extremely high learning rates (104 times higher than SRCNN ) enabled by adjustable gradient clipping. ECBSR \cite{ecbsr} proposes a reparameterizable building block called edge-guided convolutional block (ECB) for efficient SR design.
LapSRN \cite{lap} takes coarse-resolution feature maps as input, predicts high-frequency residuals, and upsamples to a finer level using transposed convolutions. No bicubic interpolation is required as a preprocessing step, thus greatly reducing computational complexity. SPAN \cite{wan2024swift} is an efficient SISR model that adopts a parameter-independent attention mechanism to improve feature extraction efficiency through symmetric activation functions (enhancing high-contribution information such as edges and textures) and residual connections (slowing down gradient disappearance and improving feature propagation).

\begin{figure*}[!t]
    \centering
   
\includegraphics[width=1\linewidth]{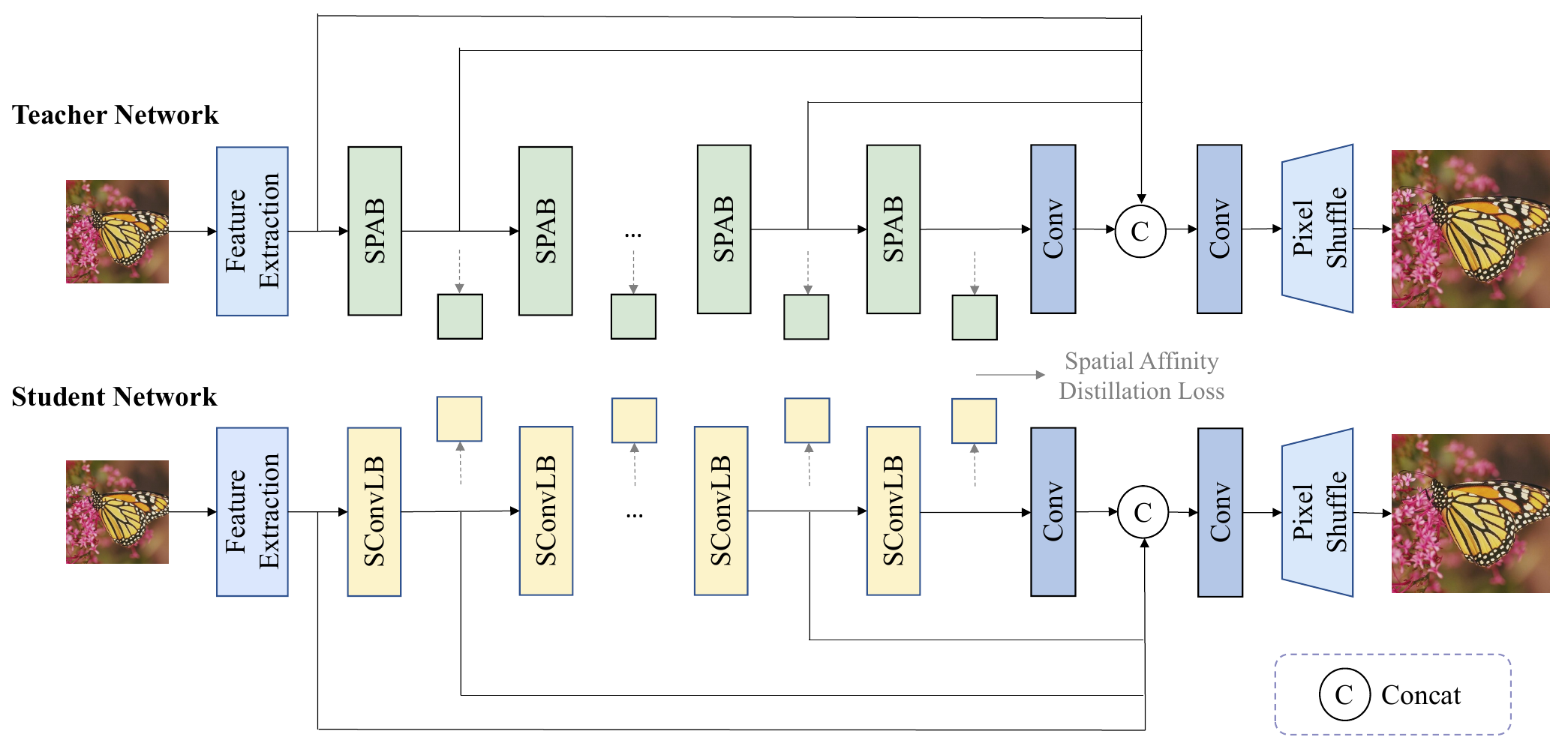} 
    \caption{The whole architecture of  DSCLoRA model. 
    We replace the SPAB module with the proposed SConvLB module and incorporate ConvLoRA layers into both the pixel shuffle block and its preceding convolutional layer.
    Spatial Affinity Distillation Loss is calculated between each feature map.}
     \label{fig:dscf}
\end{figure*}

\subsection{Distillation Network}

In the efficient SR task, the small model often needs to improve the image quality indicators (such as PSNR, SSIM) with as few parameters as possible.
Through knowledge distillation, the small model can better capture edge details, texture information and complex features, thereby effectively improving the reconstruction quality of the image.
Siqi Sun \emph{et al.} \cite{distillation1} propose a Patient Knowledge Distillation approach to compress an original large model (teacher) into an equally-effective lightweight shallow network (student). Their student model patiently learns from multiple intermediate layers of the teacher model for incremental knowledge extraction.
Wonpyo Park 
\emph{et al.} \cite{distillation2} introduce a novel approach, dubbed relational knowledge distillation (RKD), that transfers mutual relations of data examples instead. 
For concrete realizations of RKD, they propose distance-wise and angle-wise distillation losses that penalize structural differences in relations.
Feature Affinity-based Knowledge Distillation (FAKD) \cite{he2020fakd} distills the second-order statistical information from feature maps to transfer the structural knowledge of a heavy teacher model to a lightweight student model and demonstrates its efficacy over other knowledge based methods.

\subsection{Low-Rank Adaptation}
LoRA (Low-Rank Adaptation) \cite{lora1} is an efficient model fine-tuning method that aims to adapt large pre-trained models to meet the needs of specific tasks by introducing low-rank matrices. This method injects a trainable low-rank decomposition matrix into each layer of the pre-trained model without updating all model parameters, thereby significantly reducing the number of parameters and computing resources required for fine-tuning. In addition, compared with traditional full-scale fine-tuning methods, LoRA performs on par or better model quality, with higher training throughput and no increase in inference latency. This approach has been validated on models such as RoBERTa, DeBERTa, GPT-2, and GPT-3 \cite{lora1}. SoRA \cite{lora2} is a sparse low-rank adaptation method that can dynamically adjust the intrinsic rank during the adaptation process. DyLoRA \cite{lora3} method trains LoRA blocks for a range of ranks instead of a single rank by sorting the representation learned by the adapter module at different ranks during training. ConvLoRA \cite{aleem2024convlora} freezes pre-trained model weights, adds trainable low-rank decomposition matrices to convolutional layers, and backpropagates the gradient through these matrices thus greatly reducing the number of trainable parameters.


%% file: sec/3_method.tex
\section{Method}
\label{sec:3}

\subsection{Preliminary}
Our model, the DSCLoRA model, is based on the SPAN \cite{wan2024swift}. SPAN takes a low-resolution image $I_{LR}$ as input and generates a high-resolution image $I_{HR}$. 
Among them, the Swift Parameter-free Attention Block (SPAB) module is an important part of SPAN, which is shown in \cref{fig:spab}.
SPAB implements a parameter-free attention mechanism, and its structure is as follows \cite{wan2024swift}:  

\begin{figure}[t]
  \centering
\includegraphics[width=1\linewidth]{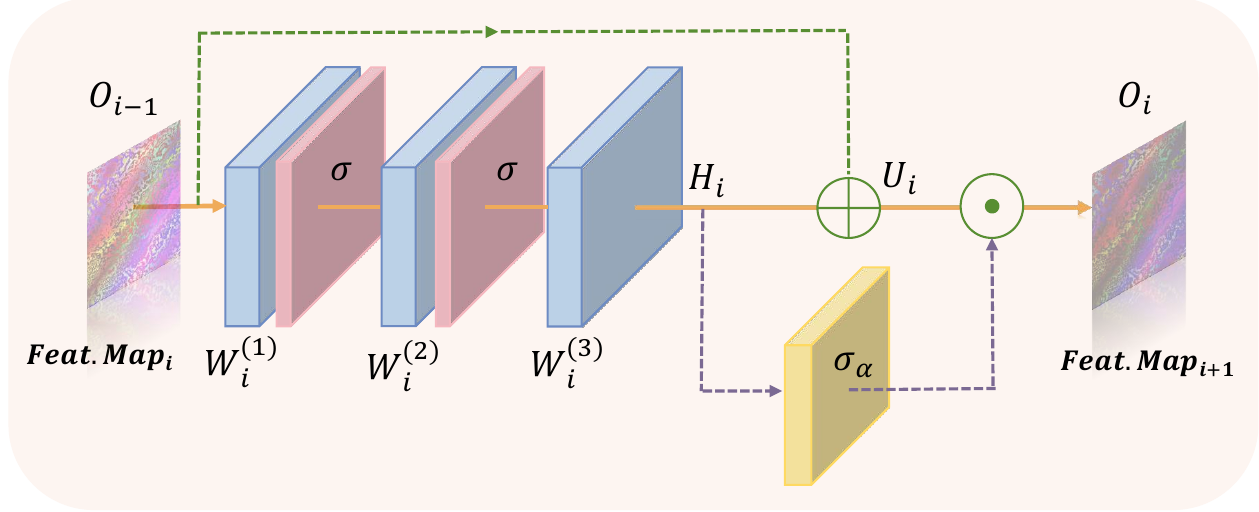}
   \caption{Schematic diagram of SPAB structure \cite{wan2024swift}.}
   \label{fig:spab}
\end{figure}

\noindent\textbf{Convolutional Feature Extraction}:  
    The input feature $O_{i-1}$ is passed through three convolutional layers to generate the feature $H_i$:  
    \begin{align}
    H_i &= F^{(i)}_{c, W_i}(O_{i-1}) \nonumber \\
        &= W_i^{(3)} \otimes \sigma\left( W_i^{(2)} \otimes \sigma\left( W_i^{(1)} \otimes O_{i-1} \right) \right),
\end{align}
    where $W_i^{(1)}, W_i^{(2)}, W_i^{(3)}$ represent the convolution kernels in the $i$-th SPAB.

\noindent\textbf{Residual Connection}:  
    The input feature $O_{i-1}$ and the convolutional feature $H_i$ are added element-wise to form the pre-attention feature map $U_i$:  
    \begin{equation}
        U_i = O_{i-1} \oplus H_i,
    \end{equation}  
    where $\oplus$ denotes element-wise addition.

\noindent\textbf{Parameter-free Attention Mechanism}:  
    The convolutional feature $H_i$ is passed through a symmetric activation function $\sigma_a(\cdot)$ to generate the attention map $V_i$:  
    \begin{equation}
        V_i = \sigma_a(H_i).
    \end{equation}  
    Then, the output feature $O_i$ is obtained by element-wise multiplication of the pre-attention feature map and the attention map:  
    \begin{equation}
        O_i = U_i \odot V_i,
    \end{equation}  
    where $\odot$ denotes element-wise multiplication.

\subsection{Network Architecture of DSCLoRA Model}
\label{subsec:network arch}

The network structure of the DSCLoRA model is shown in \cref{fig:dscf}, which is based on the leading efficient super-resolution method SPAN \cite{wan2024swift}. 

\begin{figure}[t]
  \centering
\includegraphics[width=0.8\linewidth]{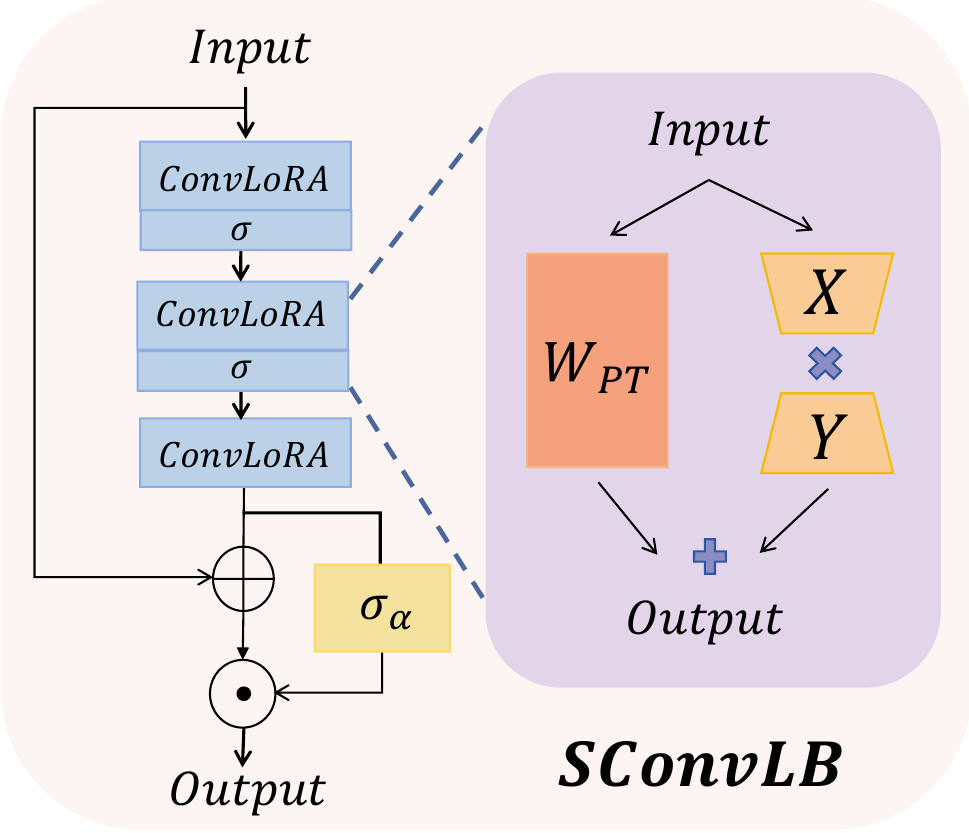}
   \caption{Schematic diagram of SConvLB structure. The $\sigma$ in the figure represents the SiLU activation layer. The pretrained model weights (denoted as $W_{PT}$) remain fixed during training, while only the LoRA parameters ($X$ and $Y$) are updated.}
   \label{fig:sconvlb}
\end{figure}

Inspired by ConvLoRA \cite{aleem2024convlora}, we propose SConvLB module, shown in \cref{fig:sconvlb}, which incorporates ConvLoRA into SPAB to improve performance without increasing computation complexity. Specifically, given a pre-trained convolutional layer in SPAB, we update it by adding LoRA layers, and representing it with a low-rank decomposition:
\begin{equation}
    W_{ConvLoRA} = W_{PT}+XY,
\end{equation}
where $W_{ConvLoRA}$ denotes the updated weight parameters of the convolution, $W_{PT}$ denotes the original pre-trained parameters of the convolution, $X$ is initialized by random Gaussian distribution, and $Y$ is zero in the beginning of training. Note that the LoRA weights can be merged into the main backbone. Therefore, our SConvLB module don't introduce extra computation during inference.

 We replace the SPAB in SPAN with our proposed SConvLB, and also add ConvLoRA into the pixel shuffle block and the convolution before it to get our final DSCLoRA model. During training, we freeze the original weight and bias of the convolution and only update the LoRA parameters.

\subsection{Knowledge Distillation Training Strategy}
\label{subsec:distillation strategy}

\begin{table*}[!t]
    \centering
    \small
    \setlength{\tabcolsep}{2.5pt} 
    \begin{tabular}{cccccccc}  
        \hline
        \multirow{2}{*}{Model} & \multirow{2}{*}{Params (K)} & \multirow{2}{*}{FLOPs (G)} & Set5 & Set14 & BSD100 & Urban100 & Manga109 \\  
    
        & & & PSNR$\uparrow$ / SSIM$\uparrow$ & PSNR$\uparrow$ / SSIM$\uparrow$ & PSNR$\uparrow$ / SSIM$\uparrow$ & PSNR$\uparrow$ / SSIM$\uparrow$ & PSNR$\uparrow$ / SSIM$\uparrow$ \\  
        \hline
        SRCNN\cite{srcnn} & 57 & 3.75 & 30.48 / 0.8628 & 27.49 / 0.7503 & 26.90 / 0.7101 & 24.52 / 0.7221 & 27.58 / 0.8555 \\  
        FSRCNN\cite{fsrcnn} & 13 & 0.82 & 30.72 / 0.8660 & 27.61 / 0.7550 & 26.98 / 0.7150 & 24.62 / 0.7280 & 27.90 / 0.8610 \\  
        VDSR\cite{vdsr} & 666 & 43.56 & 31.35 / 0.8838 & 28.01 / 0.7674 & 27.29 / 0.7251 & 25.18 / 0.7524 & 28.83 / 0.8870 \\   
        IMDN\cite{imdn} & 715 & 46.54 & 32.21 / 0.8948 & 28.58 / 0.7811 & 27.56 / 0.7353 & 26.04 / 0.7838 & 30.45 / 0.9075 \\  
        RFDN\cite{rfdn} & 550 & / & \textcolor{blue}{32.24} / 0.8952 & 28.61 / 0.7819 & 27.57 / 0.7360 & 26.11 / 0.7858 & \textcolor{blue}{30.58} / \textcolor{blue}{0.9089} \\  
        EFDN\cite{efdn} & 276 & 16.70 & 32.05 / 0.8937 & 28.58 / 0.7812 & 27.56 / 0.7353 & 26.00 / 0.7814 & 30.32 / 0.9058\\  
        SAFMN++\cite{safmn} & 212 & 13.86 & 32.01 / 0.8923 & 28.56 / 0.7805 & 27.55 / 0.7349 & 26.02 / 0.7818 & 30.21 / 0.9035 \\  
        LKDN-S\cite{lkdn} & 172 & 11.14 & 32.09 / 0.8943 & 28.60 / \textcolor{blue}{0.7822} & \textcolor{blue}{27.60} / \textcolor{blue}{0.7370} & 26.07 / 0.7844 & 30.50 / 0.9077 \\  
        DIPNet\_ntire\cite{yu2023dipnet} & 243 & 14.89 & 31.79 / 0.8903 & 28.43 / 0.7775 & 27.46 / 0.7318 & 25.76 / 0.7751 & 29.82 / 0.8993 \\  
        SPAN-Tiny\cite{wan2024swift} & 131 & 8.54 & 31.73 / 0.8896 & 28.32 / 0.7756 & 27.40 / 0.7297 & 25.55 / 0.7685 & 29.56 / 0.8967 \\ 
        RLFN\cite{rlfn} & 543 & 33.96 & \textcolor{red}{32.24} / \textcolor{blue}{0.8952} & \textcolor{blue}{28.62} / 0.7813 & 27.60 / 0.7364 & \textcolor{red}{26.17} / \textcolor{red}{0.7877} & / \\
        DSCLoRA(ours)  & 131 & 8.54 & 31.73 / 0.8897 & 28.32 / 0.7758 & 27.41 / 0.7299 & 25.58 / 0.7690 & 29.60 / 0.8971 \\  
        DSCLoRA-L(ours)  & 426 & 27.87 & 32.19 / \textcolor{red}{0.8958} & \textcolor{red}{28.63} / \textcolor{red}{0.7830} & \textcolor{red}{27.61} / \textcolor{red}{0.7370} & \textcolor{blue}{26.13} / \textcolor{blue}{0.7864} & \textcolor{red}{30.60} / \textcolor{red}{0.9095} \\  
        \hline
    \end{tabular}
    \caption{Quantitative results of the state-of-the-art ESR models on five benchmark datasets in $\times 4$ scale factor task. Parameters and FLOPs are measured with $256 \times 256$ shaped inputs. The best and second-best results are marked in \textcolor{red}{red} and \textcolor{blue}{blue} colors, respectively. }
    \label{tab:x4_results}
\end{table*}

To supervise the optimization of SConvLB, we adopt a knowledge-based distillation training strategy. We adopt spatial affinity-based knowledge distillation \cite{he2020fakd} to transfer second-order statistical info from the teacher model to the student model by aligning spatial feature affinity matrices at multiple layers of the networks. Given a feature $F_l \in R^{B\times C\times W\times H}$ extracted from the $l$-th layer of the network, we first flatten the tensor along the last two dimensions and calculate the affinity matrix $A_{spatial}$. Then the spatial feature affinity-based distillation loss can be formulated as:
\begin{equation}
    L_{AD}=\frac{1}{|A|}\sum_{l=1}^{n} ||A_l^S-A_l^T||_1,
\end{equation}
where $A_l^S$ and $A_l^T$ are the spatial affinity matrix of student and teacher networks extracted from the feature maps of the $l$-th layer, respectively. $|A|$ denotes the number of elements in the affinity matrix. Specifically, we apply the distillation loss after each SConvLB. 

Except for the distillation loss in the feature space, we apply a pixel-level distillation loss:
\begin{equation}
    L_{TS}=||\mathcal{T}(I_{LR})-\mathcal{S}(I_{LR})||_1,
\end{equation}
where $\mathcal{T}$ and $\mathcal{S}$ denote the teacher network and the student network, respectively. $I_{LR}$ denotes the LR image. 

We also apply the $L_2$ loss:
\begin{equation}
    L_{rec}=||I_{HR}-\mathcal{S}(I_{LR})||_2,
    \label{eq:rec loss}
\end{equation}
where $I_{HR}$ denotes the ground truth high-resolution image.
The overall loss is:
\begin{equation}
    L_{total} = \lambda_1L_{rec}+\lambda_2L_{TS}+\lambda_3L_{AD},
    \label{eq: total loss}
\end{equation}
where $\lambda_1$, $\lambda_2$, $\lambda_3$ are hyper-parameters.


%% file: sec/4_Experiments.tex
\section{Experiments}
\label{sec:4}
\subsection{Datasets and Metrics}
The DSCLoRA model is trained on 800 high-quality images from the DIV2K \cite{div2k} dataset and 84,991 high-quality images from the LSDIR \cite{lsdir} dataset, totaling 85,791 images.
For evaluation, We use two common metrics called peak signal-to-noise ratio (PSNR) and structure similarity index (SSIM). All images are cropped by 4 pixels along the borders to calculate PSNR and SSIM. In the DIV2K \cite{div2k} and LSDIR \cite{lsdir} dataset, PSNR and SSIM are evaluated on the RGB channels, consistent with the test protocol for the NTIRE 2025 Efficient SR competition \cite{ren2025tenth}. However, for the five benchmark datasets, including Set5 \cite{set5}, Set14 \cite{set14}, BSD100 \cite{bsd100}, Urban100 \cite{urban100}, and Manga109 \cite{manga109}, PSNR and SSIM are computed on the luminance (Y) channel. In addition, we calculate the number of parameters, FLOPs and the average inference time per image to evaluate the efficiency of our methods.

\begin{table}[t]
    \centering
    \setlength{\tabcolsep}{24pt}
    \begin{tabular}{c|cc}  
        \toprule
        layer name & $r$ & $\alpha$ \\  
        \midrule
        $SConvLB_6$ & 4 & 8 \\ 
        $SConvLB_1$ & 8 & 16 \\ 
        $SConvLB_{2,3,4,5}$ & 12 & 24 \\
        $Upsampler$ & 16 & 32 \\
        $Conv_2$ & 24 & 48 \\
        \bottomrule
    \end{tabular}
    \caption{Implementation details of DSCLoRA. $r$ and $\alpha$ denote two hyper parameters of LoRA.}
    \label{tab:impl of lora}
    \vspace{-1em}
\end{table}

\subsection{Implementation Details}
\label{subsec: implementation details}
We retain the original SPAN network structure and keep all parameters from the pre-trained weights frozen.
Then, we add LoRA and set the size of $r$ and $\alpha$ as in \cref{tab:impl of lora}. During training, we only update the LoRA parameters. The three hyper-parameters $\lambda_1$, $\lambda_2$ and $\lambda_3$ of \cref{eq: total loss} are set to 1, 0.1 and 0.01, respectively. The training process is divided into two stages: 

\noindent\textbf{Stage One}: HR patches of size $192 \times 192$ are randomly cropped from HR images, and the mini-batch size is set to 8. The model is trained by minimizing the $L_{total}$ mentioned in \cref{eq: total loss} with the Adam optimizer \cite{adam}. The learning rate is set to $1 \times 10^{-4}$. A total of $ 30k $ iterations are trained.
    
\noindent\textbf{Stage Two}: In the second stage, we increase the size of the HR image patches to $256 \times 256$, with other settings remaining the same as in the first stage.

Following the methodology proposed in \cite{ren2024ninth}, we employ the Exponential Moving Average (EMA) strategy to enhance training robustness. Random flipping and random rotation are also used for data augmentation. After the two-stage training process, we merge the LoRA layer weights with the frozen weights and perform a weighted average with the original pre-trained weights to obtain the final model. To facilitate subsequent comparisons with other methods, we trained two versions: the 26-channel DSCLoRA model and the 48-channel DSCLoRA-L model, which were distilled with the 28-channel and 52-channel SPAN \cite{wan2024swift} models, respectively.

\begin{table}[!t]
    \centering
   \small
    \setlength{\tabcolsep}{1.2pt} 
    \begin{tabular}{cccc}
        \toprule
        \multirow{2}{*}{Benchmark} & SPAN-Tiny\cite{wan2024swift} & DSCLoRA$_{\text{stage1}}$ & DSCLoRA (ours) \\  
        & PSNR / SSIM & PSNR / SSIM & PSNR / SSIM \\
        \midrule
        Set5 & 31.7252 / 0.8896 & \textcolor{blue}{31.7312} / \textcolor{blue}{0.8896} & \textcolor{red}{31.7332} / \textcolor{red}{0.8897} \\
        Set14 & 28.3151 / 0.7756 & \textcolor{blue}{28.3184} / \textcolor{blue}{0.7757} & \textcolor{red}{28.3246} / \textcolor{red}{0.7758} \\  
        BSD100 & 27.3994 / 0.7297 & \textcolor{blue}{27.4011} / \textcolor{blue}{0.7297} & \textcolor{red}{27.4057} / \textcolor{red}{0.7299} \\  
        Urban100 & 25.5544 / 0.7685 & \textcolor{blue}{25.5608} / \textcolor{blue}{0.7686} & \textcolor{red}{25.5774} / \textcolor{red}{0.7690} \\ 
        Manga109 & 29.5629 / 0.8967 & \textcolor{blue}{29.5766} / \textcolor{blue}{0.8969} & \textcolor{red}{29.5998} / \textcolor{red}{0.8971} \\  
        \bottomrule
    \end{tabular}
    \caption{Performance comparison of DSCLoRA and SPAN-Tiny (26 channels) on 5 benchmarks, with performance changes across two training phases. The best and second-best results are marked in \textcolor{red}{red} and \textcolor{blue}{blue} colors, respectively.}
    \label{tab:dscf vs span}
\end{table}

\begin{figure*}[t]
  \centering
\includegraphics[width=0.95\textwidth]{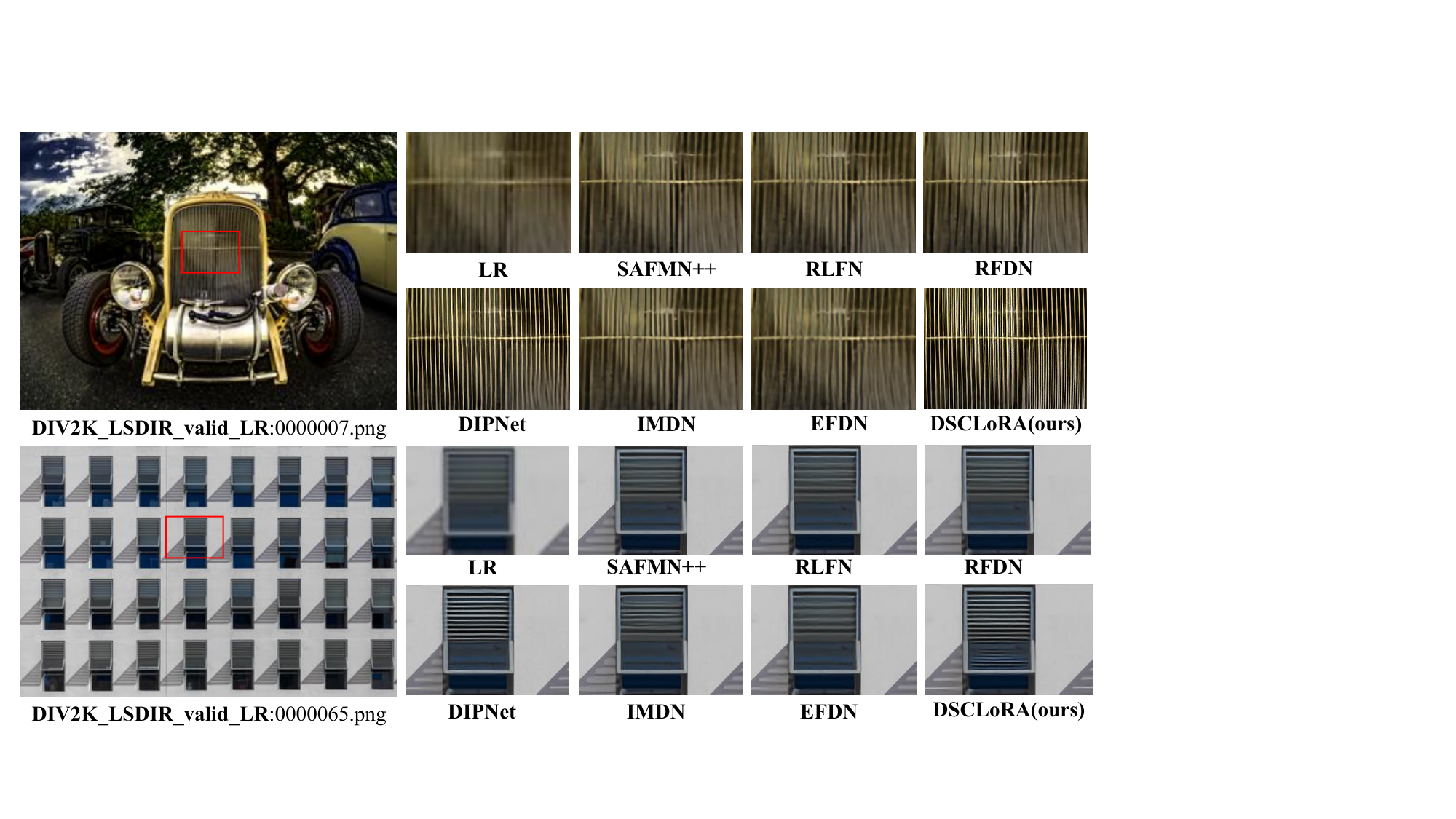}
        \caption{Visual comparison of images with rich texture details generated by different models on the DIV2K\_LSDIR\_valid dataset \cite{ren2025tenth}.}
   \label{texture}
\end{figure*}

\begin{table*}[!htbp]
    \centering
    \setlength{\tabcolsep}{8pt}
    \begin{tabular}{l|cccccc}  
        \toprule
        Model & PSNR(dB) & Time (ms) & Params (M) & FLOPS (G) & Activations (M) & Memory (M)  \\  
        \midrule
        IMDN\cite{imdn} & 26.91 & 39.41 & 0.715 & 46.54 & 122.68 & 1023.73 \\
        E-RFDN\cite{rfdn} & \textcolor{red}{27.00} & 25.61 & 0.433 & 27.05 & 112.03 & 788.13 \\
        RLFN\_ntire\cite{rlfn} & 26.96 & 29.48 & 0.317 & 19.67 & 80.05 & \textcolor{blue}{467.70} \\
        EFDN\cite{efdn} & 26.93 & 29.77 & 0.276 & 16.70 & 111.12 & 662.89 \\
        DIPNet\_ntire\cite{yu2023dipnet} & \textcolor{blue}{27.00} & 18.52 & 0.243 & 14.89 & 72.97 & 493.54 \\
        SAFMN++\cite{safmn} & 26.90 & 17.70 & 0.212 & 13.86 & 69.28 & \textcolor{red}{405.74} \\
        LKDN-S\cite{lkdn} & 26.97 & 41.59 & 0.172 & 11.14 & 247.14 & 831.84 \\
        SPAN\cite{wan2024swift} & 26.94 & \textcolor{blue}{9.06} & \textcolor{blue}{0.151} & \textcolor{blue}{9.83} & \textcolor{blue}{41.68} & 829.53 \\
        DSCLoRA(Ours) & 26.92 & \textcolor{red}{8.82} & \textcolor{red}{0.131} & \textcolor{red}{8.54} & \textcolor{red}{38.93} & 728.41 \\
        \bottomrule
    \end{tabular}
    \caption{Comparison with the complexity of some recent models in $\times 4$ scale factor task. PSNR is measured on the RGB channel, and the entire metric is based on the DIV2K\_LSDIR\_valid dataset \cite{ren2025tenth} with an NVIDIA RTX A6000 GPU. The "Time" metric represents the average duration spent reasoning about each image. For each model, the reasoning time was measured five times, and the average of these measurements was calculated. FLOPs are measured with $256 \times 256$ shaped inputs. "SPAN" in the table is the 28-channel version. The best and second-best results are marked in \textcolor{red}{red} and \textcolor{blue}{blue} colors, respectively. }
    \label{tab:comparison}
\end{table*}

\begin{figure*}[t]
  \centering
\includegraphics[width=0.95\textwidth]{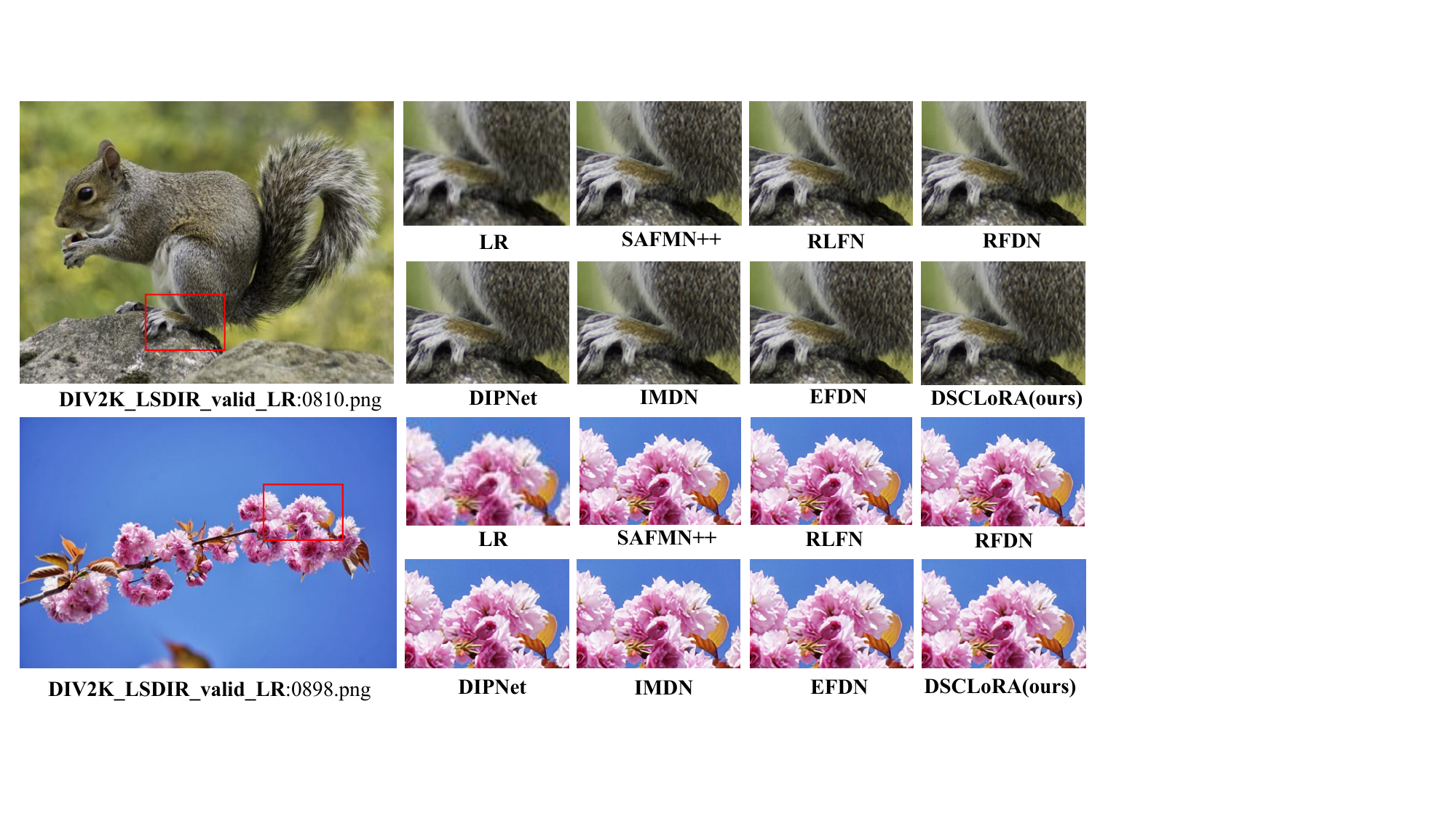}
        \caption{Visual comparison of images with rich natural details generated by different models on the DIV2K\_LSDIR\_valid dataset \cite{ren2025tenth}.}
        \label{other}
\end{figure*}

\subsection{Quantitative Comparison}
In this study, we have done superscoring work mainly on $\times 4$ upsampling on a variety of benchmarks, and then compared our approach with a variety of existing state-of-the-art efficient superscoring models \cite{srcnn,fsrcnn,vdsr,drcn,imdn,rfdn,rlfn,efdn,safmn,lkdn,yu2023dipnet,wan2024swift}. As seen in \cref{tab:dscf vs span}, our method gives better PSNR and SSIM on multiple benchmarks than the method before fine-tuning with LoRA, which shows the effectiveness of our method and progressive training strategy.
Larger scale comparisons need to be attended to in \cref{tab:x4_results}, which shows the PSNR and SSIM of many classical methods on 5 benchmarks. As can be seen from \cref{tab:x4_results}, the 48-channel DSCLoRA-L model performs optimally or sub-optimally on almost all benchmarks while keeping the number of parameters small. And even the lightweight DSCLoRA model with only 26 channels has quite good PSNR and SSIM on all benchmarks.

\subsection{Qualitative Comparison}

As shown in \cref{texture} and \cref{other}, our model demonstrates good performance in both texture and natural detail recovery. In \cref{texture}, our model significantly outperforms others in preserving fine texture details, which are crucial for high-quality image reconstruction. This result is particularly important for applications where preserving intricate textures, such as fabric patterns or surface structures, is vital. 
In \cref{other}, it can be seen that in terms of natural details, even though our model has improved the computing speed, it has not sacrificed the performance of detail processing. DSCLoRA maximizes performance while improving efficiency.


\begin{table}[t]
    \centering
    \footnotesize
    \setlength{\tabcolsep}{2.5pt} 
    \renewcommand{\arraystretch}{1} 
    \begin{tabular}{cccc}  
        \toprule
        \multirow{2}{*}{Teacher Model} & None & SPAN\_ch52 & SPAN\_ch28 (ours) \\  
        & PSNR / SSIM & PSNR / SSIM & PSNR / SSIM \\
        \midrule
        Set5 & \textcolor{blue}{31.7285} / \textcolor{blue}{0.8896} & 31.7268 / 0.8896 & \textcolor{red}{31.7332} / \textcolor{red}{0.8897} \\
        Set14 & \textcolor{blue}{28.3195} / \textcolor{blue}{0.7757} & 28.3159 / 0.7757 & \textcolor{red}{28.3246} / \textcolor{red}{0.7758} \\  
        BSD100 & \textcolor{blue}{27.4023} / \textcolor{blue}{0.7298} & 27.4003 / 0.7297 & \textcolor{red}{27.4057} / \textcolor{red}{0.7299} \\  
        Urban100 & \textcolor{blue}{25.5656} / \textcolor{blue}{0.7687} & 25.5566 / 0.7686 & \textcolor{red}{25.5774} / \textcolor{red}{0.7690} \\ 
        Manga109 & \textcolor{blue}{29.5744} / \textcolor{blue}{0.8968} & 29.5677 / 0.8968 & \textcolor{red}{29.5998} / \textcolor{red}{0.8971} \\          
        \bottomrule
    \end{tabular}
    \caption{Performance comparison of DSCLoRA on five benchmarks trained without distillation, with SPAN\_ch52 and SPAN\_ch28 as teacher models, respectively. The best and second-best results are marked in \textcolor{red}{red} and \textcolor{blue}{blue} colors, respectively.}
    \label{tab:teacher ablation}
\end{table}

\subsection{Model Complexity}

As can be seen from the \cref{tab:comparison}, our model has a very low number of parameters and FLOPs while keeping the PSNR comparable to other models. Notably, DSCLoRA model maintains the fastest inference time and the advantage of consuming fewer resources of the original lightweight model. This is due to the fine-tuning of the convolutional layers while keeping the original lightweight model structure unchanged by the DSCLoRA method , which leads to a further improvement in the performance of the model.


\begin{table}[t]
    \centering
    \footnotesize 
    \setlength{\tabcolsep}{4pt} 
    \begin{tabular}{cccc}
        \toprule
        \multirow{2}{*}{Benchmark} & w/o SA & $L1$ & DSCLoRA (ours) \\  
        & PSNR / SSIM & PSNR / SSIM & PSNR / SSIM \\
        \midrule
        Set5 & \textcolor{blue}{31.7269} / \textcolor{blue}{0.8896} & 31.7264 / 0.8896 & \textcolor{red}{31.7332} / \textcolor{red}{0.8897} \\
        Set14 & 28.3159 / \textcolor{blue}{0.7757} & \textcolor{blue}{28.3161} / 0.7756 & \textcolor{red}{28.3246} / \textcolor{red}{0.7758} \\  
        BSD100 & \textcolor{blue}{27.4003} / \textcolor{blue}{0.7297} & 27.3998 / 0.7297 & \textcolor{red}{27.4057} / \textcolor{red}{0.7299} \\  
        Urban100 & 25.5566 / \textcolor{blue}{0.7686} & \textcolor{blue}{25.5567} / 0.7685 & \textcolor{red}{25.5774} / \textcolor{red}{0.7690} \\  
        Manga109 & \textcolor{blue}{29.5677} / \textcolor{blue}{0.8968} & 29.5670 / 0.8968 & \textcolor{red}{29.5998} / \textcolor{red}{0.8971} \\ 
        \bottomrule
    \end{tabular}
    \caption{Results of the ablation study focusing on the impact of different distillation loss functions across five benchmarks. The best and second-best results are marked in \textcolor{red}{red} and \textcolor{blue}{blue} colors, respectively.}
    \label{tab:loss ablation}
\end{table}

\subsection{Ablation Studies}
In the ablation experiment, we focus on $\times 4$ super-resolution task.
Unless otherwise specified, the training configuration is consistent with the implementation details described in \cref{subsec: implementation details}.

\noindent\textbf{Selection of Teacher Model.} To explore the effect of teacher model choice on model performance, the 52-channel SPAN model and the 28-channel SPAN model are used as teacher models to instruct the 26-channel DSCLoRA model to learn respectively and the results obtained are shown in \cref{tab:teacher ablation}. We also present the results without distillation in \cref{tab:teacher ablation}.



The results in \cref{tab:teacher ablation} indicate that distilling from a large-channel model to a small-channel model under the same training configuration performs worse than using a medium-channel model as the teacher, and is even slightly inferior to no distillation. This may be due to the mismatch in spatial affinity between the feature layers of large-channel model and small-channel model, which hampers effective knowledge transfer.

\begin{table}[t]
    \centering
    \scriptsize 
    \setlength{\tabcolsep}{1.3pt} 
    \begin{tabular}{ccccc}
        \toprule
        \multirow{2}{*}{Benchmark} & w/o LoRA & part LoRA & LoRA-r2 & DSCLoRA (ours) \\  
        & PSNR / SSIM & PSNR / SSIM & PSNR / SSIM & PSNR / SSIM \\
        \midrule
        Set5 & 31.7241 / 0.8895 & \textcolor{blue}{31.7279} / \textcolor{blue}{0.8896} & 31.7257 / 0.8896 & \textcolor{red}{31.7332} / \textcolor{red}{0.8897} \\
        Set14 & 28.3173 / 0.7756 & \textcolor{blue}{28.3183} / \textcolor{blue}{0.7756} & 28.3157 / 0.7756 & \textcolor{red}{28.3246} / \textcolor{red}{0.7758} \\  
        BSD100 & 27.3993 / 0.7296 & 27.3996 / 0.7297 & \textcolor{blue}{27.3999} / \textcolor{blue}{0.7297} & \textcolor{red}{27.4057} / \textcolor{red}{0.7299} \\  
        Urban100 & 25.5558 / 0.7684 & \textcolor{blue}{25.5563} / \textcolor{blue}{0.7685} & 25.5561 / 0.7685 & \textcolor{red}{25.5774} / \textcolor{red}{0.7690} \\  
        Manga109 & 29.5656 / 0.8967 & \textcolor{blue}{29.5749} / \textcolor{blue}{0.8968} & 29.5653 / 0.8968 & \textcolor{red}{29.5998} / \textcolor{red}{0.8971} \\  
        \bottomrule
    \end{tabular}
    \caption{Results of the ablation study focusing on the impact of LoRA across five benchmarks. The best and second-best results are marked in \textcolor{red}{red} and \textcolor{blue}{blue} colors, respectively.}
    \label{tab:convlora ablation}
\end{table}

\noindent\textbf{Loss Function of Distillation.} 
Spatial affinity-based (SA) knowledge distillation is employed in DSCLoRA training. To validate the effectiveness of SA, we replace the intermediate feature layer loss function. Since SPAN (28 channels) guides DSCLoRA (26 channels), we use $1 \times 1$ convolutions to align DSCLoRA's feature channels to 28, enabling pixel-level L1 loss computation:  
$ L_\text{feat} = \frac{1}{n} \sum_{l=1}^{n} \| F_l^T - \psi(F_l^S) \|_1 $,  
where $F_l^T$ and $F_l^S$ are the $l$-th layer feature maps of teacher and student networks, and $\psi$ denotes $1 \times 1$ convolution for channel alignment. Minimizing the total loss:  
$ L_{total} = \lambda_1L_{rec}+\lambda_2L_{TS}+\lambda_3L_{feat} $  
produces the "w/o SA" model. To assess contribution of $L_2$ loss, we further replace the $L_2$ loss in \cref{eq:rec loss} with $L_1$ loss, resulting in the "$L1$" model. Both models are trained under identical settings as DSCLoRA and evaluated on benchmark datasets, with results presented in \cref{tab:loss ablation}.

\noindent\textbf{LoRA in DSCLoRA.} 
To investigate the impact of adding LoRA to convolutional layers for fine-tuning, including which layers to apply LoRA and the optimal $r$ value, we conduct the following experiments: (1) "w/o LoRA" model—LoRA removed, all weights unfrozen, trained under the same configuration; (2) "part LoRA" model—LoRA applied only to $conv_2$ and $upsampler$ layers; (3) "LoRA-r2" model—$r=2$, $\alpha=4$ for all LoRA layers. Results are shown in \cref{tab:convlora ablation}.


\cref{tab:convlora ablation} demonstrates that incorporating LoRA layers leads to notable performance gains while direct full fine-tuning without LoRA is less effective under the current configuration. Further observation reveals that fine-tuning more convolution layers with appropriately high $r$ values yields significant PSNR and SSIM improvements.

We selected some state-of-the-art LoRA variants \cite{LoRA+,LoRAFA,rsLoRA} for comparative analysis by replacing the LoRA type in DSCLoRA. With the patch size set to $ 256 \times 256 $ directly, learning rate to $10^{-5}$, and 500 training iterations, other configurations remained unchanged. \cref{tab:lora type ablation} shows minimal performance differences across LoRA types, with standard LoRA achieving satisfactory results.
We did not adopt the visual adapter approach, as it requires additional layers and overhead, contrary to our original intent.

\begin{table}[t]
    \centering
    \scriptsize 
    \setlength{\tabcolsep}{1.3pt} 
    \begin{tabular}{ccccc}
        \toprule
        \multirow{2}{*}{LoRA Type} & LoRA+\cite{LoRA+} & LoRA-FA\cite{LoRAFA} & rsLoRA\cite{rsLoRA} & LoRA (ours) \\  
        & PSNR / SSIM & PSNR / SSIM & PSNR / SSIM & PSNR / SSIM \\
        \midrule
        Set5 & \textcolor{red}{31.7273} / \textcolor{red}{0.8896} & 31.7251 / 0.8896 & 31.7256 / 0.8896 & \textcolor{blue}{31.7268} / \textcolor{blue}{0.8896} \\
        Set14 & \textcolor{blue}{28.3162} / \textcolor{blue}{0.7756} & 28.3151 / 0.7756 & 28.3154 / 0.7756 & \textcolor{red}{28.3168} / \textcolor{red}{0.7756} \\  
        BSD100 & \textcolor{red}{27.3997} / \textcolor{red}{0.7297} & 27.3994 / 0.7297 & 27.3994 / 0.7297 & \textcolor{blue}{27.3996} / \textcolor{blue}{0.7297} \\  
        Urban100 & \textcolor{blue}{25.5567} / \textcolor{blue}{0.7685} & 25.5543 / 0.7685 & 25.5549 / 0.7685 & \textcolor{red}{25.5572} / \textcolor{red}{0.7685} \\  
        Manga109 & \textcolor{blue}{29.5663} / \textcolor{blue}{0.8968} & 29.5629 / 0.8967 & 29.5638 / 0.8967 & \textcolor{red}{29.5672} / \textcolor{red}{0.8968} \\  
        \bottomrule
    \end{tabular}
    \caption{Performance comparison of DSCLoRA on 5 different benchmarks using LoRA+, LoRA-FA, rsLoRA and LoRA training respectively. The best and second-best results are marked in \textcolor{red}{red} and \textcolor{blue}{blue} colors, respectively.}
    \label{tab:lora type ablation}
\end{table}

\subsection{DSCLoRA for NTIRE 2025 Challenge}
We participated in the NTIRE 2025 Efficient Super-Resolution Challenge \cite{ren2025tenth}, which aimed to optimize trade-offs between runtime, model complexity, and computational efficiency. 
We optimized the SPAN-Tiny model \cite{wan2024swift} using the DSCLoRA method proposed in this paper, resulting in the development of the DSCF model. The DSCF model successfully achieved the PSNR thresholds without introducing additional computational costs. Consequently, the DSCF model secured 1st place in the Overall Performance Track of the competition.




%% file: sec/5_Conclusion.tex
\section{Conclusion}
\label{sec:5}
In this paper, we propose Distillation-Supervised Convolutional Low-Rank Adaptation (DSCLoRA) to enhance the performance of existing efficient super-resolution (SR) models. By introducing ConvLoRA \cite{aleem2024convlora} and employing spatial affinity-based knowledge distillation \cite{he2020fakd} to align feature affinity matrices across multiple network layers, DSCLoRA effectively transfers second-order statistical information from teacher to student models. Built upon the SPAN model \cite{wan2024swift}, our method enhances pre-trained efficient SR models without incurring additional computational cost and requires only a limited amount of training to achieve substantial improvements in PSNR and SSIM metrics. Experimental results across multiple benchmarks demonstrate that DSCLoRA consistently outperforms the original SPAN model and achieves superior trade-offs between model size and performance compared to other state-of-the-art efficient SR methods. These results highlight the effectiveness of our approach, and future work may explore applying DSCLoRA to a broader range of models to further enhance their performance.

%% file: sec/6_acknowledgment.tex
\section*{Acknowledgment}
\label{sec:6}
This work was partly supported by Science and Technology Commission of Shanghai Municipality (No.24511106200), the Shanghai Key Laboratory of Digital Media Processing and Transmission under Grant 22DZ2229005, 111 project BP0719010.